\pdfoutput=1

\documentclass[11pt]{article}
\usepackage[]{emnlp2021}

\usepackage{times}
\usepackage{latexsym}

\usepackage[T1]{fontenc}

\usepackage[utf8]{inputenc}

\usepackage{microtype}
\usepackage{graphicx}
\usepackage{subfigure}
\usepackage{ulem}
\usepackage{caption}
\usepackage{CJKutf8}
\usepackage{multirow}
\usepackage{amsmath}
\title{Emily: Developing An Emotion-affective Open-Domain Chatbot with Knowledge Graph-based Persona}

\newcommand*{\affaddr}[1]{#1} 
\newcommand*{\affmark}[1][*]{\textsuperscript{#1}}
\newcommand*{\email}[1]{\texttt{#1}}

\author{%
Weixuan Wang\affmark[1] \footnotemark[1] , Xiaoling Cai\affmark[1] \footnotemark[1], Chong Hsuan Huang, \\\textbf{Haoran Wang, Haonan Lu, Ximing Liu\affmark[1], Wei Peng\affmark[1]\footnotemark[2]}\\
\affaddr{\affmark[1]Artificial Intelligence Application Research Center, Huawei Technologies}\\
\email{\{peng.wei1\}@huawei.com}\\%
}

\date{}

\begin{document}
\begin{CJK}{UTF8}{gbsn}
\maketitle

\renewcommand{\thefootnote}{\fnsymbol{footnote}}
\footnotetext[1]{Co-first authors.}
\footnotetext[2]{Corresponding author.}
\renewcommand{\thefootnote}{\arabic{footnote}}

\begin{abstract}
In this paper, we describe approaches for developing Emily, an emotion-affective open-domain chatbot. Emily can perceive a user's negative emotion state and offer supports by positively converting the user's emotion states. This is done by finetuning a pretrained dialogue model upon data capturing dialogue contexts and desirable emotion states transition across turns. Emily can differentiate a general open-domain dialogue utterance with questions relating to personal information. By leveraging a question-answering approach based on knowledge graphs to handle personal information, Emily maintains personality consistency. We evaluate Emily against a few state-of-the-art open-domain chatbots and show the effects of the proposed approaches in emotion affecting and addressing personality inconsistency.    
\end{abstract}

\section{Introduction}
Developing dialogue systems capable of responding to human emotions has emerged as a major research stream to enhance human engagements in conversation. These research works focused on developing conversational agents perceiving and expressing emotions, for example, ``empathetic listeners/chatbots'' \citep{moel,caire}, and ``emotional chatting machines'' \citep{emotional}. While the studies mentioned above focused on generating empathetic responses from the system side, another stream of research works relating to emotion elicitation commenced exploring the effects of the agent's responses on users' emotion states \citep{example,elicit,emoelicit}. A concurrent research work addressing negative human emotions in dialog systems has recently coined the notion of ``Emotional Support Conversation (ESC)'' \citep{huang}, in which a task framework of ESC is defined. In this paper, we describe approaches for developing Emily, an emotion-affective open-domain chatbot built by finetuning a pretrained dialogue model upon data capturing dialogue contexts and desirable emotion states transition across turns.        

Another major issue associated with open-domain chatbots is the lack of consistent personality \citep{vinyals2015neural, li2016persona}. Despite the ongoing efforts in incorporating persona information into dialogue generation, the issue remains unsolved due to a lack of control over the decoding process  \citep{li2016persona, zhang2018personalizing,song2020profile}. In order to reduce the discrepancy between personalized information and the responses from a chatbot, we replace dialogue generation with a retrieval-based approach when addressing personality-related questions. We leverage a question-answering (QA) approach based on knowledge graphs (KGQA) to retrieve personal information embedded as attribute graphs. Emily first classifies users' utterances to identify the personalized questions which is subsequently handled by the KGQA model resulting in more accurate responses. In addition, Emily is equipped with a world celebrity knowledge graph to offer answers relating to celebrities.

Besides presenting the way developing Emily, which is an emotion-affective open-domain chatbot, the contributions of this work are as follows:

\begin{itemize}
\item We produce a dataset allowing for modeling users' positive emotion state transitions;
\item We further develop a KGQA approach to address personality inconsistency between personal information and the responses; 
\item We evaluate Emily against a few state-of-the-art (SOTA) open-domain chatbots and show the effects of proposed approaches in emotion affecting and providing personality consistency.
\end{itemize}

\begin{figure*}[!h]
    \centering
    \includegraphics[width=0.9\textwidth,height=0.3\textheight]{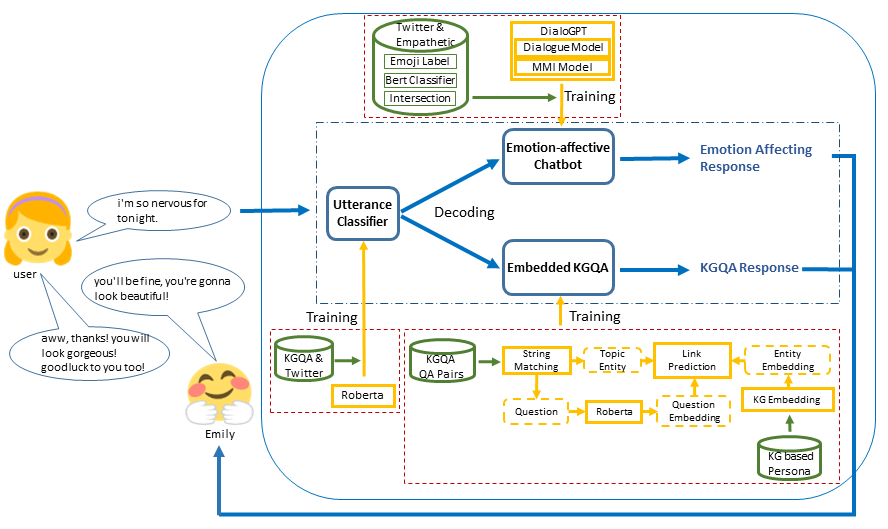}
    \caption{The training and decoding processes of Emily. The part in green presents the data augmentation. The modules in yellow solid shape are the pretrained models (Roberta, DialoGPT). The parts in blue (Utterance Classifier, Emotion-affective Chatbot, and Embedded KGQA) are the final constitutes of Emily used for decoding. The modules enclosed in the red dot line represent the training phrase, and those in the blue dash-dot line indicating the decoding phrase.}
    \label{framework}
\end{figure*}

\section{Methodology}
Figure~\ref{framework} shows the way Emily is constructed. Emily consists of three key components: the utterance classifier, the emotion-affective module, and the Embedded KGQA persona module, depicted in the following subsections.     
\subsection{Utterance Classifier}
The utterance classifier is designed to identify a persona-related question that the Embedded KGQA persona module handles to ensure consistency. The emotion-affective module deals with the general non-persona-related utterances. The classifier is built by finetuning the pretrained language model ``Roberta'' \citep{roberta} on the data with statistics depicted in Table~\ref{classifier-data}. 

\begin{table}
\centering
\begin{tabular}{l|c|c}
\hline
~ & \textbf{Persona-related} & \textbf{General}\\ \hline
\textbf{Training Data} & 150,000  & 150,000 \\ \hline
\textbf{Validation Data} & 10,000 & 10,000 \\  \hline
\textbf{Test Data} & 10,000 & 10,000 \\ \hline
\end{tabular}
\caption{\label{classifier-data} The data used to train the utterance classifier. }
\end{table}

The probability is computed as follows during inference:
\begin{equation}
P(z|x)=softmax(WQ)
\end{equation}
where $Q$ is the vector representation of the question encoded by ``Roberta'' and the $W$ is the parameter of the classifier. The model uses the cross-entropy loss function to calculate the classification loss. The utterance classifier can achieve an accuracy of 99.5\%.

\subsection{Emotion-affective Modelling}

\subsubsection{Data Preparation}
\label{dataprep}

We collect data from Twitter and Empathetic data set available from \citet{emoelicit} and \citet{benchmark}. We remove the symbol ``@user'', quotation marks, and hashtags from the Twitter dataset to normalize the data. As for the Twitter dataset, 58 common emojis are used to label the emotion states (positive, neutral, negative), consistent with the selection rules mentioned in ``MOJITALK'' \citep{mojitalk}. When multiple types of emojis appear in the dialogue sentence, we choose the most frequent emoji to represent the emotion state of the sentence. 

It is critical to classify the Twitter data to produce valid emotion labels. The following steps are performed:

\begin{enumerate}
\item We train an emotion classifier for states (positive, neutral, and negative) by finetuning BERT \citep{bert} on the Google Play Store Apps review data available from Kaggle \footnote{https://www.kaggle.com/lava18/google-play-store-apps};
\item The classifier is used to identify the emotion states of each dialogue sentence; 
\item The sentence is also tagged with the above-mentioned emoji labeling method;
\item The intersection of the classification result and the emoji label is used to produce the final result of the emotion state of the sentence.
\end{enumerate}

As for the empathetic dataset, we also map 32 evenly distributed emotion labels for the empathetic dataset into the three categories mentioned above. 
We subsequently perform a data filtering process in which only dialogue turns with positive users' emotion state transitions are selected from the two data sources mentioned above.

\subsubsection{Finetuning Pretrained Dialogue Response Generation Model}

The emotion-affective chatbot is developed based on the DialoGPT-large architecture \citep{dialogpt}, initialized with DialoGPT-large model parameters of 762M. The base model is finetuned with the filtered data mentioned in Section~\ref{dataprep} until it converges. In addition to the model parameters, we also learn the word embeddings of the tokens with a size of 1,280. The size of the vocabulary is 50,257.


The maximum mutual information (MMI) score function \citep{dialogpt} is implemented to regulate generation in removing uninformative responses.  


\subsection{Developing Embedded KGQA-based Persona}

\subsubsection{Embedded KGQA}
The KGQA module is based on the work of Embedded KGQA \citep{saxena2020improving}.  
The Knowledge Graph $\mathbf{G}$ consists of a series of triples, each represented as $(h,r,t)$ that indicates a relation $r$ from a head entity $h$ to a tail entity $t$. Knowledge Graph Embedding (KGE) targets at learning a low-dimensional vector representation for entities and relations denoted as $\mathbf{E}$ and $\mathbf{R}$ respectively and the $i^{th}$ entity and $j^{th}$ relationship are denoted as $E_{i}$ and $R_{j}$. ComplEx \citep{trouillon2016complex} embeds entities and relations in complex space and defines a scoring function as:
\begin{equation} 
    \begin{aligned}
        \phi(h,r,t)&=Re(<E_{h},R_{r},\bar{E_{t}}>) \\
        & =Re(\sum_{k=1}^{d}E_{hk},R_{rk},\bar{E_{tk}}) \\
        & =<Re(E_{h}),Re(R_{r}),Re(E_{t})> \\
        & +<Re(E_{h}),Im(R_{r}),Im(E_{t})> \\
        & +<Im(E_{h}),Re(R_{r}),Im(E_{t})> \\
        & -<Im(E_{h}),Im(R_{r}),Re(E_{t})>
    \end{aligned}
\end{equation}\label{eqn2}
$E_{h}\in C^{d},R_{r}\in C^{d},\bar{E_{t}}\in C^{d}$ represent the head entity vector, the relation vector and the conjugate of the tail entity vector respectively, while $d$ represents the dimension of a vector. $Re()$ is a function to obtain the real part of a complex vector and $Im()$ extracts the imaginary part. For a score of a true triplet in KG greater than 0, it can be represented as $\phi(h,r,t)>0$. $\phi(h,r,\bar{t})<0$ is used to describe a negative triplet which a tail entity $t$ is randomly replaced by another entity $\bar{t}$. 
As shown in Figure~\ref{framework}, for each natural language question $q$, we use the maximum string matching method to obtain the topic entity and map it to the representation $E_{h}\in C^{d}$. And the model embeds the question to a 768-dimensional vector by ``Roberta'' and then converts the vector to a fixed dimension vector $Q_{q}\in C^{d}$ by a feed-forward neural network. While the question is embedded, the answer $a\in A$ is mapped to a fixed dimension vector $E_{a}\in C^{d_e}$ where $A$ are the answer entities set of question.

\begin{equation}
\phi(h,q,a)=\left\{
\begin{array}{rcl}
Re({< E_{h},Q_{q},E_{a}>})>0,a\in A \\
Re({< E_{h},Q_{q},E_{a}>})<0,a\not \in A
\end{array}
\right.
\end{equation}

At the training stage, the scoring function is calculated with all the entities in KG to produce the predicted score distributions. Kullback-Leibler divergence \citep{kl} loss function is used to measure the distance from the predicted distributions to the true answer distributions.

\subsubsection{Data Preparation}
\label{data}
\paragraph{Knowledge Graph-based Persona:}We construct a knowledge graph about Emily along with some celebrities whose information are collected from the website \footnote{https://www.thefamouspeople.com/}. As shown in Table~\ref{persona-table}, the persona information contains a number of personality settings (i.e., ``Name'', ``Birthday''). 

\begin{table}
\centering
\begin{tabular}{ll}
\hline \textbf{Personalized Key} & \textbf{Personalized Value} \\ \hline
Name & Emily \\
Birthday & Privacy \\
Nationality & World \\
Age & Forever Young \\
Gender & Privacy \\
Born Country & China \\
Famous as & AI\\
City & shenzhen \\
\hline
\end{tabular}
\caption{\label{persona-table} Persona information for Emily. }
\end{table}

\paragraph{Question Answering Dataset:}For training the QA model, we build a dataset based on the knowledge graph. Following the way that MetaQA \citep{zhang2017variational} is generated, we design about 10 question templates for each relationship in KG and perform stratified random sampling from them when generating questions. For example, we design the question templates such as ``How old are the $NE$'' and ``Tell me $NE$'s age'' for the relation ``Age''. The ``$NE$'' can be replaced by a specific person's name, while the age value would be the answer. To increase the robustness of the 
data, we apply the back-translation \citep{back} approach using a pretrained neural machine translation model for data enhancement. Entities are guaranteed to be kept in the paraphrased question. The statistics of the dataset are shown in Table~\ref{kgqa-data}.


\begin{table}[]
\begin{tabular}{c|c|c|c}
\hline
\textbf{KG-based} & \textbf{Triplet} &  \textbf{Entity} & \textbf{Relation} \\ \cline{2-4} 
\textbf{Persona}  & 187,116 & 59,672     & 23       \\ \hline
\textbf{\multirow{2}{*}{QA pairs}} & \textbf{Train}   & \textbf{Validation} & \textbf{Test}     \\ \cline{2-4} 
& 143,776 & 20,000     & 20,000   \\ \hline
\end{tabular}
\caption{\label{kgqa-data} The data used in Embedded KGQA Module. }
\end{table}

\section{Experiments and Evaluation}
In this section, we describe a range of experiments comparing Emily with two SOTA open-domain chatbots, Blender \citep{blender} and PLATO-2 \citep{plato}, on related metrics. It should be bear in mind that Emily and other open-domain chatbots are designed for different purposes; the results only indicate the effectiveness of Emily in emotion affecting and maintaining personality consistency.

\subsection{Automatic Evaluation}

\subsubsection{Evaluation Metrics}

We adopt three well-accepted automatic evaluation metrics to compare Emily with Human (reference data), Blender and PLATO-2.

\begin{itemize}
\item \textbf{Context: } We measure the context correlation between the responses produced by the dialogue model and the queries, similar to a metric used in \citet{metric}.

\item \textbf{Fluency: }Fluency is often measured using a language model. It indicates the negative perplexity of generated responses. We adopt the calculation method proposed by \citet{metric} to produce the fluency score.

\end{itemize}


\subsubsection{Automatic Evaluation Results}

We randomly sample 100 dialogues from an empathetic dialogue test dataset \citep{benchmark}. The corresponding response generations are produced by Emily, Blender, and PLATO-2. Then we visualize the distribution of results on each evaluation metric. 

\begin{figure*}[htbp]
\centering
\subfigure[Distributions of the context metric of responses recorded for human, Emily, Blender and PLATO-2.]{
\includegraphics[width=7cm]{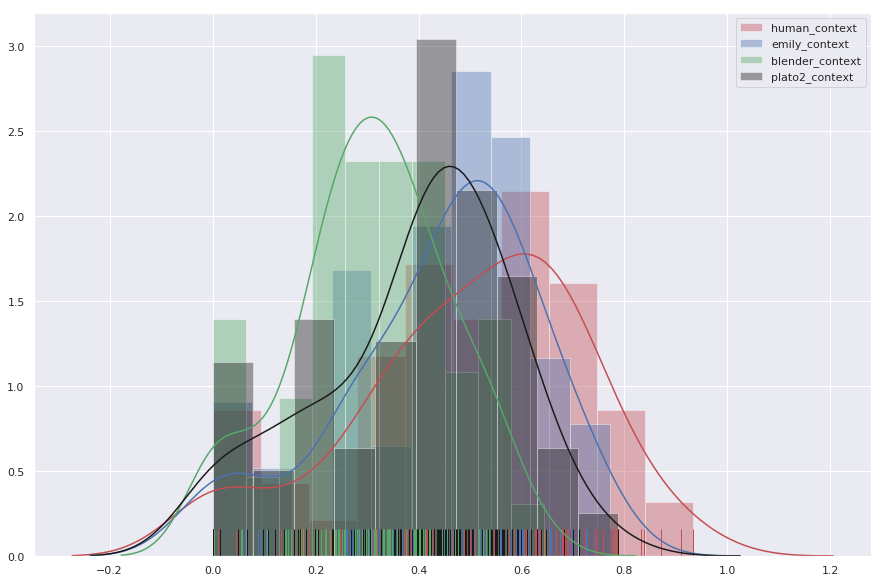}
\label{fig1}
}
\quad
\subfigure[Distributions of the fluency metric of responses recorded for human, Emily, Blender and PLATO-2.]{
\includegraphics[width=7cm]{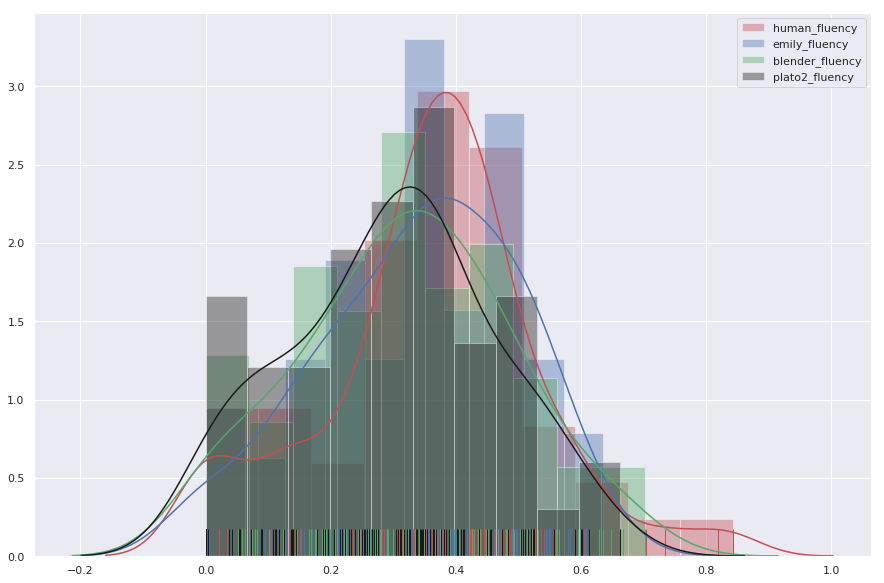}
\label{fig2}
}
\caption{Distribution results on the automatic evaluation metrics.}
\label{all}
\end{figure*}

The distribution of context is shown in Figure~\ref{fig1}. It is clearly shown that Emily (the blue histogram/distribution) achieves a higher context correlation with human judgments (the red histogram/distribution) than those recorded for Blender and PLATO-2. In addition, Figure~\ref{fig2} demonstrates that Emily achieves competitive results in comparison to those of Blender and PLATO-2.


\subsection{Human Evaluation Results}

To further measure the emotion-affective effect of the generated responses, we conduct human evaluations on empathetic dialogue test dataset \citep{benchmark} for Human (reference data), Emily, and Blender. Following an approach mentioned in \citet{moel}, we randomly select 100 dialogues and their corresponding responses. We ask a human evaluator to blind review the systems mentioned above on three metrics: Empathy, Relevance, and Fluency, all rated on a Likert scale of 1 to 5 (1: not at all, 3: somewhat, 5: very much):

\begin{itemize}
\item \textbf{Empathy: } The metric of empathy is used to measure whether the system can perceive users' emotion states and to elicit positive emotions subsequently;

\item \textbf{Relevance: } We ask a human evaluator to judge whether the responses are relevant to the user's topics; 

\item \textbf{Fluency: } We utilize fluency to estimate whether the human evaluator can understand the system-generated responses and whether the responses are correct in grammar.
\end{itemize}

\begin{table}
\centering
\begin{tabular}{llll}
\hline 
\textbf{Model} & \textbf{Empathy} & \textbf{Relevance} & \textbf{Fluency} \\ 
\hline
\textbf{Human} & 4.05 & 4.51 & 4.76 \\
\textbf{Emily} & 3.93 & 4.23 & 4.72  \\
\textbf{Blender} & 3.96 & 3.99 & 4.84 \\
\hline
\end{tabular}
\caption{\label{human-table} Human evaluation results are mean scores from blind-reviewing the generated responses for Emily, Blender, and Human (the reference data).}
\end{table}



The evaluation results are shown in Table~\ref{human-table}. Emily achieves an equivalent empathy score to those of human reference data and Blender. It can be observed from Table~\ref{human-table} that Emily scores higher than Blender in the relevance metric. It is interesting to find out Blender is able to produce a fluency score higher than that of reference data, probably due to the prevalence of texting-induced unconventional English in the reference data.  

\begin{table}
\centering
\begin{tabular}{l|l}
\hline \textbf{Question} & \textbf{Answer} \\ \hline
May I know your name please? & Emily \\
Could you tell me your name? & Emily \\
Hello! What's your name? & Emily \\
your name, please & Emily \\
Would you mind telling me your name? & Emily \\
May I make bold to ask your name? & Emily \\
How shall I call you? & Emily \\
I like to know your name please & Emily \\
How shall I address you? & Emily \\
\hline
\end{tabular}
\caption{\label{sample1} Building variations of persona-related questions into Emily. }
\end{table}

\subsection{Embedded KGQA Evaluation}

We describe the performance of Emily in addressing persona-related questions via Embedded KGQA \citep{saxena2020improving}. As mentioned before, the personal information and celebrity-related knowledge are organized as knowledge graphs. The entities in the graph are the answers to the questions. For this task, we use hit@1, hit@3, and hit@10 to evaluate the performance of the model, as shown in Table~\ref{KGQA-Evaluation}.

\begin{table}
\centering
\begin{tabular}{c|c|c|c}
\hline ~ & \textbf{hit@1} & \textbf{hit@3} & \textbf{hit@10} \\
\hline
\textbf{Embedded KGQA} & 0.9778 & 0.9898 & 0.9925   \\
\hline
\end{tabular}
\caption{\label{KGQA-Evaluation} Evaluation of the Embedded KGQA on personal information relating to Emily and celebrities. }
\end{table}


The model has achieved a high hit@1 addressing the questions of personal information for Emily and celebrities. Some examples of the persona collected are presented in Table~\ref{sample1}. The variations of personal questions are built into the knowledge graphs and handled by the Embedded KGQA module. 

\section{Related Work}
\subsection{Emotion-affective Dialog System}
Emotion-aware chatbot has become an emerging area of research in recent years. \citet{emotional} first addressed the emotion factor in large-scale dialog generation using an end-to-end framework to generate contents and emotions. \citet{mojitalk} proposed a sophisticated CVAE-based model called ``MOJITALK'', which used emoji to control the emotion and sentiment of the generated responses. \citet{caire} presented an empathetic chatbot ``CAiRE'' which fine-tunes a large-scale pretrained language model with multiple objectives aiming at detecting dialogue emotion and generating empathetic responses. \citet{benchmark} focused on empathetic dialogue generation and released a novel empathetic dialogue dataset as a benchmark. \citet{happy} trained a sentiment predictor with a reinforcement learning framework to encourage more empathetic responses. \citet{moel} introduced a novel dialogue system named ``MoEL'' to perceive the users' feelings and respond accordingly by learning specific listeners for each emotion state.

 
 \citet{example} utilized many examples of human appraisal in spoken dialogue to elicit a positive emotional impact throughout the interaction. \citet{elicit} built a chat-oriented dialogue system that can dynamically mimic affective human interaction and generate more natural responses aiming at eliciting a more positive emotional impact. \citet{emoelicit} proposed a variational model EmoElicitor to generate responses that can elicit users' specific emotions with the help of a pretrained language model. \citet{huang} defined a task framework for developing ESC to reduce users' emotional distress via a three-stage procedure (exploration, comforting, and action) and supporting strategies. 


\subsection{Personality Consistency}
The first attempt to model persona can be seen in \citet{li2016persona} where a speaker model is used to capture individual characteristics. \citet{qian2017assigning} designed a model to decide whether a post should be responded to based on pre-specified agent profiles. To solve the user-sparsity problem and the issue w.r.t. meaningless responses \citep{zhang2018personalizing,song2020profile}, new persona models and high-quality data have been introduced. \citet{zhang2018personalizing} introduced a Persona-Chat dataset and proposed two generative models to handle persona-related information. \citet{zheng2019personalized} contributed a multi-turn dialogue dataset containing various traits from a large number of speakers. A human-annotated dataset with single-turn conversations and key-value attribute profiles were created by \citep{song2020profile}.

To address factoid questions based on incomplete KG, \citet{sun2018open,sun2019pullnet} drew answers from topic entity-related corpora and from knowledge-based sources. \citet{huang2019knowledge} and \citet{saxena2020improving} leveraged the knowledge graph embedding (KGE) to deal with the problem of sparse graphs. KGE is used to learn a low-dimensional vector representation for each entity and relation to preserve the original structure, for example in TransE \citep{bordes2013translating}, ComplEx \citep{trouillon2016complex}. Unlike \citet{huang2019knowledge} focusing on simple questions, \citet{saxena2020improving} proposed an ``EmbedKGQA'' model performing multi-hop KGQA over sparse KG.  

\section{Conclusion}
In this work, we present approaches for developing an emotion-affective open-domain chatbot Emily. Emily is designed to generate emotion affecting responses in response to a negative emotion state. This is performed by finetuning a pretrained dialogue model upon data capturing dialogue contexts and desirable user emotion states transition. Leveraging an utterance classifier and Embedded KGQA module, Emily can handle persona-related questions in a consistent way. Experimental results demonstrate Emily's effectiveness in emotion affecting and addressing personality inconsistency. Future work will focus on enhancing the emotion affecting capability and benchmark the results in more extensive experiment settings.

\bibliography{anthology,custom}

\begin{thebibliography}{31}
\expandafter\ifx\csname natexlab\endcsname\relax\def\natexlab#1{#1}\fi

\bibitem[{Bao et~al.(2021)Bao, He, Wang, Wu, Wang, Wu, Guo, Liu, and
  Xu}]{plato}
Siqi Bao, Huang He, Fan Wang, Hua Wu, Haifeng Wang, Wenquan Wu, Zhen Guo,
  Zhibin Liu, and Xinchao Xu. 2021.
\newblock \href {https://aclanthology.org/2021.findings-acl.222} {{PLATO-2:}
  towards building an open-domain chatbot via curriculum learning}.
\newblock In \emph{Findings of the Association for Computational Linguistics:
  {ACL/IJCNLP} 2021, Online Event, August 1-6, 2021}, pages 2513--2525.
  Association for Computational Linguistics.

\bibitem[{Bordes et~al.(2013)Bordes, Usunier, Garcia-Duran, Weston, and
  Yakhnenko}]{bordes2013translating}
Antoine Bordes, Nicolas Usunier, Alberto Garcia-Duran, Jason Weston, and Oksana
  Yakhnenko. 2013.
\newblock Translating embeddings for modeling multi-relational data.
\newblock \emph{Advances in neural information processing systems},
  26:2787--2795.

\bibitem[{Devlin et~al.(2019)Devlin, Chang, Lee, and Toutanova}]{bert}
Jacob Devlin, Ming-Wei Chang, Kenton Lee, and Kristina Toutanova. 2019.
\newblock Bert: Pre-training of deep bidirectional transformers for language
  understanding.
\newblock In \emph{NAACL-HLT}.

\bibitem[{Huang et~al.(2019)Huang, Zhang, Li, and Li}]{huang2019knowledge}
Xiao Huang, Jingyuan Zhang, Dingcheng Li, and Ping Li. 2019.
\newblock Knowledge graph embedding based question answering.
\newblock In \emph{Proceedings of the Twelfth ACM International Conference on
  Web Search and Data Mining}, pages 105--113.

\bibitem[{Kullback and Leibler(1951)}]{kl}
Solomon Kullback and Richard~A Leibler. 1951.
\newblock On information and sufficiency.
\newblock \emph{The annals of mathematical statistics}, 22(1):79--86.

\bibitem[{Li et~al.(2016)Li, Galley, Brockett, Spithourakis, Gao, and
  Dolan}]{li2016persona}
Jiwei Li, Michel Galley, Chris Brockett, Georgios~P Spithourakis, Jianfeng Gao,
  and Bill Dolan. 2016.
\newblock A persona-based neural conversation model.
\newblock \emph{arXiv preprint arXiv:1603.06155}.

\bibitem[{Li et~al.(2020)Li, Feng, Wang, Song, Zhang, and Wang}]{emoelicit}
Shifeng Li, Shi Feng, Daling Wang, Kaisong Song, Yifei Zhang, and Weichao Wang.
  2020.
\newblock \href {https://doi.org/10.24963/ijcai.2020/503} {Emoelicitor: An open
  domain response generation model with user emotional reaction awareness}.
\newblock In \emph{Proceedings of the Twenty-Ninth International Joint
  Conference on Artificial Intelligence, {IJCAI} 2020}, pages 3637--3643.
  ijcai.org.

\bibitem[{Lin et~al.(2019)Lin, Madotto, Shin, Xu, and Fung}]{moel}
Zhaojiang Lin, Andrea Madotto, Jamin Shin, Peng Xu, and Pascale Fung. 2019.
\newblock \href {https://doi.org/10.18653/v1/D19-1012} {Moel: Mixture of
  empathetic listeners}.
\newblock In \emph{Proceedings of the 2019 Conference on Empirical Methods in
  Natural Language Processing and the 9th International Joint Conference on
  Natural Language Processing, {EMNLP-IJCNLP} 2019, Hong Kong, China, November
  3-7, 2019}, pages 121--132. Association for Computational Linguistics.

\bibitem[{Lin et~al.(2020)Lin, Xu, Winata, Siddique, Liu, Shin, and
  Fung}]{caire}
Zhaojiang Lin, Peng Xu, Genta~Indra Winata, Farhad~Bin Siddique, Zihan Liu,
  Jamin Shin, and Pascale Fung. 2020.
\newblock \href {https://aaai.org/ojs/index.php/AAAI/article/view/7098} {Caire:
  An end-to-end empathetic chatbot}.
\newblock In \emph{The Thirty-Fourth {AAAI} Conference on Artificial
  Intelligence, {AAAI} 2020, The Thirty-Second Innovative Applications of
  Artificial Intelligence Conference, {IAAI} 2020, The Tenth {AAAI} Symposium
  on Educational Advances in Artificial Intelligence, {EAAI} 2020, New York,
  NY, USA, February 7-12, 2020}, pages 13622--13623. {AAAI} Press.

\bibitem[{Liu et~al.(2021)Liu, Zheng, Demasi, Sabour, Li, Yu, Jiang, and
  Huang}]{huang}
Siyang Liu, Chujie Zheng, Orianna Demasi, Sahand Sabour, Yu~Li, Zhou Yu, Yong
  Jiang, and Minlie Huang. 2021.
\newblock \href {https://doi.org/10.18653/v1/2021.acl-long.269} {Towards
  emotional support dialog systems}.
\newblock In \emph{Proceedings of the 59th Annual Meeting of the Association
  for Computational Linguistics and the 11th International Joint Conference on
  Natural Language Processing, {ACL/IJCNLP} 2021, (Volume 1: Long Papers),
  Virtual Event, August 1-6, 2021}, pages 3469--3483. Association for
  Computational Linguistics.

\bibitem[{Liu et~al.(2019)Liu, Ott, Goyal, Du, Joshi, Chen, Levy, Lewis,
  Zettlemoyer, and Stoyanov}]{roberta}
Yinhan Liu, Myle Ott, Naman Goyal, Jingfei Du, Mandar Joshi, Danqi Chen, Omer
  Levy, Mike Lewis, Luke Zettlemoyer, and Veselin Stoyanov. 2019.
\newblock \href {http://arxiv.org/abs/1907.11692} {Roberta: {A} robustly
  optimized {BERT} pretraining approach}.
\newblock \emph{CoRR}, abs/1907.11692.

\bibitem[{Lubis et~al.(2018)Lubis, Sakti, Yoshino, and Nakamura}]{example}
Nurul Lubis, Sakriani Sakti, Koichiro Yoshino, and Satoshi Nakamura. 2018.
\newblock \href
  {https://www.aaai.org/ocs/index.php/AAAI/AAAI18/paper/view/16317} {Eliciting
  positive emotion through affect-sensitive dialogue response generation: {A}
  neural network approach}.
\newblock In \emph{Proceedings of the Thirty-Second {AAAI} Conference on
  Artificial Intelligence, (AAAI-18), the 30th innovative Applications of
  Artificial Intelligence (IAAI-18), and the 8th {AAAI} Symposium on
  Educational Advances in Artificial Intelligence (EAAI-18), New Orleans,
  Louisiana, USA, February 2-7, 2018}, pages 5293--5300. {AAAI} Press.

\bibitem[{Lubis et~al.(2019)Lubis, Sakti, Yoshino, and Nakamura}]{elicit}
Nurul Lubis, Sakriani Sakti, Koichiro Yoshino, and Satoshi Nakamura. 2019.
\newblock \href {https://doi.org/10.1109/TASLP.2019.2900910} {Positive emotion
  elicitation in chat-based dialogue systems}.
\newblock \emph{{IEEE} {ACM} Trans. Audio Speech Lang. Process.},
  27(4):866--877.

\bibitem[{Pang et~al.(2020)Pang, Nijkamp, Han, Zhou, Liu, and Tu}]{metric}
Bo~Pang, Erik Nijkamp, Wenjuan Han, Linqi Zhou, Yixian Liu, and Kewei Tu. 2020.
\newblock \href {https://www.aclweb.org/anthology/2020.acl-main.333/} {Towards
  holistic and automatic evaluation of open-domain dialogue generation}.
\newblock In \emph{Proceedings of the 58th Annual Meeting of the Association
  for Computational Linguistics, {ACL} 2020, Online, July 5-10, 2020}, pages
  3619--3629. Association for Computational Linguistics.

\bibitem[{Qian et~al.(2017)Qian, Huang, Zhao, Xu, and Zhu}]{qian2017assigning}
Qiao Qian, Minlie Huang, Haizhou Zhao, Jingfang Xu, and Xiaoyan Zhu. 2017.
\newblock Assigning personality/identity to a chatting machine for coherent
  conversation generation.
\newblock \emph{arXiv preprint arXiv:1706.02861}.

\bibitem[{Rashkin et~al.(2019)Rashkin, Smith, Li, and Boureau}]{benchmark}
Hannah Rashkin, Eric~Michael Smith, Margaret Li, and Y{-}Lan Boureau. 2019.
\newblock \href {https://doi.org/10.18653/v1/p19-1534} {Towards empathetic
  open-domain conversation models: {A} new benchmark and dataset}.
\newblock In \emph{Proceedings of the 57th Conference of the Association for
  Computational Linguistics, {ACL} 2019, Florence, Italy, July 28- August 2,
  2019, Volume 1: Long Papers}, pages 5370--5381. Association for Computational
  Linguistics.

\bibitem[{Roller et~al.(2020)Roller, Dinan, Goyal, Ju, Williamson, Liu, Xu,
  Ott, Shuster, Smith, Boureau, and Weston}]{blender}
Stephen Roller, Emily Dinan, Naman Goyal, Da~Ju, Mary Williamson, Yinhan Liu,
  Jing Xu, Myle Ott, Kurt Shuster, Eric~Michael Smith, Y{-}Lan Boureau, and
  Jason Weston. 2020.
\newblock \href {http://arxiv.org/abs/2004.13637} {Recipes for building an
  open-domain chatbot}.
\newblock \emph{CoRR}, abs/2004.13637.

\bibitem[{Saxena et~al.(2020)Saxena, Tripathi, and
  Talukdar}]{saxena2020improving}
Apoorv Saxena, Aditay Tripathi, and Partha Talukdar. 2020.
\newblock Improving multi-hop question answering over knowledge graphs using
  knowledge base embeddings.
\newblock In \emph{Proceedings of the 58th Annual Meeting of the Association
  for Computational Linguistics}, pages 4498--4507.

\bibitem[{Sennrich et~al.(2016)Sennrich, Haddow, and Birch}]{back}
Rico Sennrich, Barry Haddow, and Alexandra Birch. 2016.
\newblock \href {https://doi.org/10.18653/v1/p16-1009} {Improving neural
  machine translation models with monolingual data}.
\newblock In \emph{Proceedings of the 54th Annual Meeting of the Association
  for Computational Linguistics, {ACL} 2016, August 7-12, 2016, Berlin,
  Germany, Volume 1: Long Papers}. The Association for Computer Linguistics.

\bibitem[{Shin et~al.(2019)Shin, Xu, Madotto, and Fung}]{happy}
Jamin Shin, Peng Xu, Andrea Madotto, and Pascale Fung. 2019.
\newblock \href {http://arxiv.org/abs/1906.08487} {Happybot: Generating
  empathetic dialogue responses by improving user experience look-ahead}.
\newblock \emph{CoRR}, abs/1906.08487.

\bibitem[{Song et~al.(2020)Song, Wang, Zhang, Zhao, Liu, and
  Liu}]{song2020profile}
Haoyu Song, Yan Wang, Wei-Nan Zhang, Zhengyu Zhao, Ting Liu, and Xiaojiang Liu.
  2020.
\newblock Profile consistency identification for open-domain dialogue agents.
\newblock \emph{arXiv preprint arXiv:2009.09680}.

\bibitem[{Sun et~al.(2019)Sun, Bedrax-Weiss, and Cohen}]{sun2019pullnet}
Haitian Sun, Tania Bedrax-Weiss, and William~W Cohen. 2019.
\newblock Pullnet: Open domain question answering with iterative retrieval on
  knowledge bases and text.
\newblock \emph{arXiv preprint arXiv:1904.09537}.

\bibitem[{Sun et~al.(2018)Sun, Dhingra, Zaheer, Mazaitis, Salakhutdinov, and
  Cohen}]{sun2018open}
Haitian Sun, Bhuwan Dhingra, Manzil Zaheer, Kathryn Mazaitis, Ruslan
  Salakhutdinov, and William~W Cohen. 2018.
\newblock Open domain question answering using early fusion of knowledge bases
  and text.
\newblock \emph{arXiv preprint arXiv:1809.00782}.

\bibitem[{Trouillon et~al.(2016)Trouillon, Welbl, Riedel, Gaussier, and
  Bouchard}]{trouillon2016complex}
Th{\'e}o Trouillon, Johannes Welbl, Sebastian Riedel, {\'E}ric Gaussier, and
  Guillaume Bouchard. 2016.
\newblock Complex embeddings for simple link prediction.
\newblock International Conference on Machine Learning (ICML).

\bibitem[{Vinyals and Le(2015)}]{vinyals2015neural}
Oriol Vinyals and Quoc Le. 2015.
\newblock A neural conversational model.
\newblock \emph{arXiv preprint arXiv:1506.05869}.

\bibitem[{Zhang et~al.(2018)Zhang, Dinan, Urbanek, Szlam, Kiela, and
  Weston}]{zhang2018personalizing}
Saizheng Zhang, Emily Dinan, Jack Urbanek, Arthur Szlam, Douwe Kiela, and Jason
  Weston. 2018.
\newblock Personalizing dialogue agents: I have a dog, do you have pets too?
\newblock \emph{arXiv preprint arXiv:1801.07243}.

\bibitem[{Zhang et~al.(2020)Zhang, Sun, Galley, Chen, Brockett, Gao, Gao, Liu,
  and Dolan}]{dialogpt}
Yizhe Zhang, Siqi Sun, Michel Galley, Yen{-}Chun Chen, Chris Brockett, Xiang
  Gao, Jianfeng Gao, Jingjing Liu, and Bill Dolan. 2020.
\newblock \href {https://www.aclweb.org/anthology/2020.acl-demos.30/}
  {{DIALOGPT} : Large-scale generative pre-training for conversational response
  generation}.
\newblock In \emph{Proceedings of the 58th Annual Meeting of the Association
  for Computational Linguistics: System Demonstrations, {ACL} 2020, Online,
  July 5-10, 2020}, pages 270--278. Association for Computational Linguistics.

\bibitem[{Zhang et~al.(2017)Zhang, Dai, Kozareva, Smola, and
  Song}]{zhang2017variational}
Yuyu Zhang, Hanjun Dai, Zornitsa Kozareva, Alexander~J Smola, and Le~Song.
  2017.
\newblock Variational reasoning for question answering with knowledge graph.
\newblock \emph{arXiv preprint arXiv:1709.04071}.

\bibitem[{Zheng et~al.(2019)Zheng, Chen, Huang, Liu, and
  Zhu}]{zheng2019personalized}
Yinhe Zheng, Guanyi Chen, Minlie Huang, Song Liu, and Xuan Zhu. 2019.
\newblock Personalized dialogue generation with diversified traits.
\newblock \emph{arXiv preprint arXiv:1901.09672}.

\bibitem[{Zhou et~al.(2018)Zhou, Huang, Zhang, Zhu, and Liu}]{emotional}
Hao Zhou, Minlie Huang, Tianyang Zhang, Xiaoyan Zhu, and Bing Liu. 2018.
\newblock \href
  {https://www.aaai.org/ocs/index.php/AAAI/AAAI18/paper/view/16455} {Emotional
  chatting machine: Emotional conversation generation with internal and
  external memory}.
\newblock In \emph{Proceedings of the Thirty-Second {AAAI} Conference on
  Artificial Intelligence, (AAAI-18), the 30th innovative Applications of
  Artificial Intelligence (IAAI-18), and the 8th {AAAI} Symposium on
  Educational Advances in Artificial Intelligence (EAAI-18), New Orleans,
  Louisiana, USA, February 2-7, 2018}, pages 730--739. {AAAI} Press.

\bibitem[{Zhou and Wang(2018)}]{mojitalk}
Xianda Zhou and William~Yang Wang. 2018.
\newblock \href {https://www.aclweb.org/anthology/P18-1104/} {Mojitalk:
  Generating emotional responses at scale}.
\newblock In \emph{Proceedings of the 56th Annual Meeting of the Association
  for Computational Linguistics, {ACL} 2018, Melbourne, Australia, July 15-20,
  2018, Volume 1: Long Papers}, pages 1128--1137. Association for Computational
  Linguistics.

\end{thebibliography}
\bibliographystyle{acl_natbib}

\end{CJK}
\end{document}